\documentclass{article}

\usepackage[preprint]{neurips_2024}

\usepackage[utf8]{inputenc} 
\usepackage[T1]{fontenc}    
\usepackage{hyperref}       
\usepackage{url}            
\usepackage{booktabs}       
\usepackage{amsfonts}       
\usepackage{nicefrac}       
\usepackage{microtype}      
\usepackage{xcolor}         
\usepackage{graphicx} 
\usepackage{multirow}
\usepackage{array}
\usepackage{xcolor}
\usepackage{amsmath}
\usepackage{geometry}
\usepackage{caption}
\usepackage{booktabs}
\usepackage{tabularx}
\usepackage{float}
\usepackage{wrapfig}
\usepackage{authblk}

\title{LLEMamba: Low-Light Enhancement via Relighting-Guided Mamba with Deep Unfolding Network}

\author[1]{Xuanqi Zhang}
\author[2]{Haijin Zeng}
\author[3]{Jinwang Pan}
\author[1]{Qiangqiang Shen}
\author[1]{Yongyong Chen}
\affil[1]{Harbin Institute of Technology, Shenzhen}
\affil[2]{IMEC-UGent}
\affil[3]{Harbin Institute of Technology}

\begin{document}

\maketitle

\begin{abstract}
Transformer-based low-light enhancement methods have yielded promising performance by effectively capturing long-range dependencies in a global context. However, their elevated computational demand limits the scalability of multiple iterations in deep unfolding networks, and hence they have difficulty in flexibly balancing interpretability and distortion. To address this issue, we propose a novel Low-Light Enhancement method via relighting-guided Mamba with a deep unfolding network (\emph{LLEMamba}), whose theoretical interpretability and fidelity are guaranteed by Retinex optimization and Mamba deep priors, respectively. 
Specifically, our LLEMamba first constructs a Retinex model with deep priors, embedding the iterative optimization process based on the Alternating Direction Method of Multipliers (ADMM) within a deep unfolding network.
Unlike Transformer, to assist the deep unfolding framework with multiple iterations, the proposed LLEMamba introduces a novel Mamba architecture with lower computational complexity, which not only achieves light-dependent global visual context for dark images during reflectance relight but also optimizes to obtain more stable closed-form solutions.
Experiments on the benchmarks show that LLEMamba achieves superior quantitative evaluations and lower distortion visual results compared to existing state-of-the-art methods.

\end{abstract}

\section{Introduction}

Low-light image enhancement is an essential function within the domains of computer vision and image processing, which not only aids in better visual perception but also boosts the efficacy of downstream applications such as autonomous vehicles, object recognition, and semantic segmentation  \cite{li2021low, hou2024global}.
Historically, various methodologies have been employed to address the challenge of enhancing images captured under low-light conditions, including histogram equalization  \cite{pizer1987adaptive}, Retinex-based model  \cite{jobson1997multiscale}, and learning-based strategy  \cite{guo2020zero} to manipulate the luminance and contrast of such images. 

Deep learning models that utilize end-to-end convolutional neural networks (CNNs) to relight low-light images have been actively developed  \cite{zhang2022deep,wang2022low,xu2023low}. However, the performance of these learning-based methods is limited due to the lack of large-scale training data to uncover the diversity and complex mappings from low-light to normal-light conditions  \cite{ren2019low}. 
In the meantime, model-based algorithms such as Retinex-based models, formulate an input image as pixel-wise multiplication of illumination and reflectance, and attempt to estimate ideal illumination and reflectance by deriving their optimal closed-form solutions with theoretical guarantee  \cite{park2017low,li2018structure,lin2022low}.
However, on the one hand, these model-based Retinex methods reveal limitations in denoising and restoring the reflectance parts in the low-light images, due to the lack of well-designed priors. On the other hand, these methods show slow convergence, due to the inevitable hundreds of iterations and hand-crafted constraints that are difficult to optimize.

\begin{figure*}
	\centering
        \includegraphics[width=1.0\linewidth]{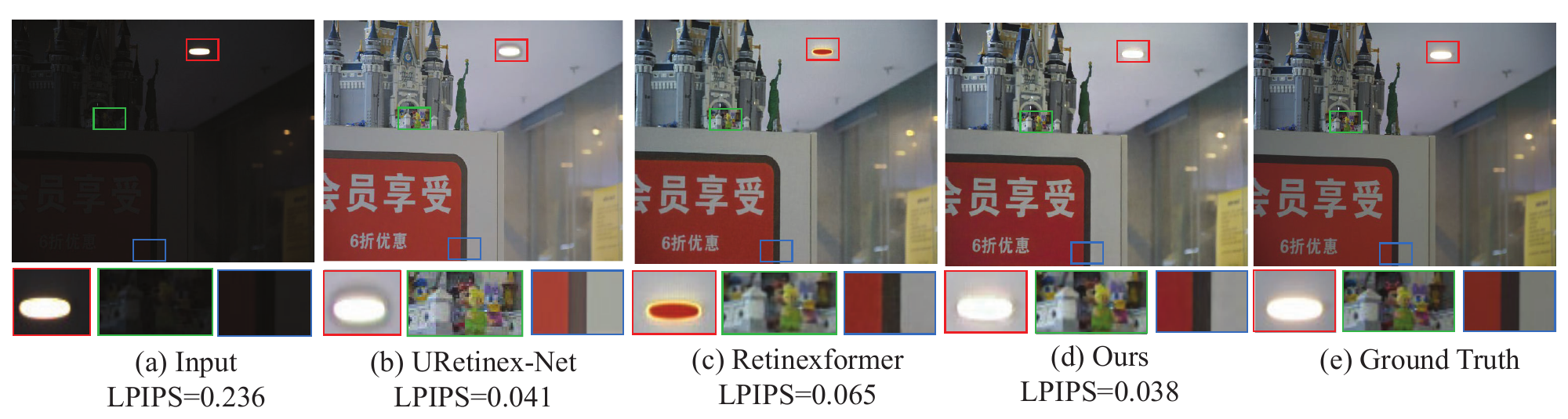}
	\caption{Example of (b) URetinex-Net \cite{wu2022uretinex}, (c) Retinexformer  \cite{cai2023retinexformer}, and the proposed (d) LLEMamba for reconstruction error (i.e., color distortion, noise, artifact) comparison in the light area. In contrast to existing state-of-the-art low-light enhancement methods URetinex-Net, Retinexformer, our LLEManba not only restores the bright areas without overexposure but also relights the dark areas without noise and color distortion. This benefits from the stability of Retinex optimization and the sustainability of long-range dependencies. }
	\label{comparison_1}  
\end{figure*}

Accordingly, Retinex-based unfolding networks are proposed to deal with the above defects of both deep learning and Retinex methods while retaining their benefits  \cite{zheng2021adaptive,wu2022uretinex,liu2023low}, which have shown promising performance and explainability in different benchmarks. Among them, Zheng et al. \cite{zheng2021adaptive} proposed to incorporate total variation regularization into the unfolding process, and could recover more details by learning a noise level map. Beyond their proposed UTVNet \cite{zheng2021adaptive}, Wu et al. \cite{wu2022uretinex} replaced the traditional prior of total variation by
two deep priors, while Li et al. \cite{liu2023low} integrated both implicit priors learned from data and explicit priors inherited from traditional methods into algorithm unrolling networks and could capture both global and local brightness.
However, these Retinex-based unfolding techniques just rely on simple convolutional neural networks as the deep priors, and hence fail to acquire the long-range dependencies and non-local self-similarity in both light and dark areas for low-light images. 
One clear example is shown in Fig. \ref{comparison_1}. For the Retinex-based unfolding model URetinex-Net  \cite{wu2022uretinex}, the exposure to the lighting causes distant objects to appear excessively white and bright, losing their original color information.

Recently, to compensate for the aforementioned deficiencies, some advanced architectures, such as vision Transformers, have been employed for better reflectance restoration  \cite{xu2022snr,cai2023retinexformer,wang2023low,ye2024codedbgt}. For example, to maintain the ability of ong-range dependencies, Cai et al. \cite{cai2023retinexformer} and Wang et al. \cite{wang2023low} presented Retinexformer and IAGC models, respectively. The former emphasized developing an Illumination-Guided Transformer after a Retinex decomposition to manage the non-local interactions across regions with varying lighting conditions, while the latter designed a new Transformer block that fully modeled pixel dependencies through a local-to-global hierarchical attention approach, enabling dark areas to leverage information from distant, informative regions effectively.
However, these Transformer-based models are often encountered with overexposure in lighting areas, as shown in Fig. \ref{comparison_1}.  Not only that, they generally bear a tedious training process and higher computational complexity that significantly hinder the expansion of Retinex-based unfolding networks into the Transformer framework, and hence face challenges in adeptly managing the trade-off between interpretability and image distortion. 

\begin{figure*}
	\centering
        \includegraphics[width=1\linewidth]{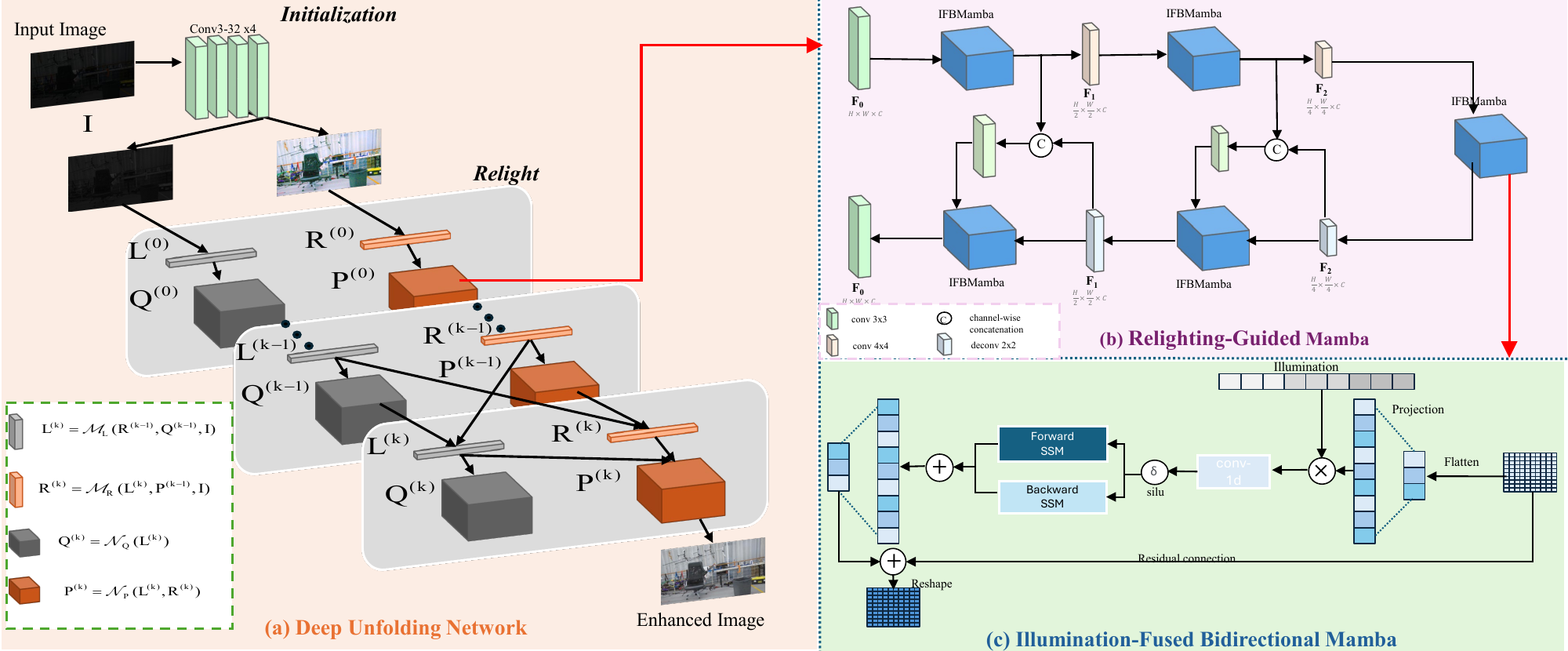}
	\caption{Overview of our proposed Relighting-guided Mamba with Deep Unfolding Network. (a) In the Deep Unfolding Network for Relight, the input is first decomposed by convolutional layers. Then decomposed images are optimized by deep unfolding layers in $k$ iterations to output the enhanced images. (b) The reflectance in $\mathbf{P}^{(k)}$ is processed by Relighting-Guided Mamba. The Relighting-Guided Mamba is a U-shaped network with Illumination-Fused Bidirectional Mamba (IFBMamba) as the backbone. (c) Then, the reflectance is processed by IFBMamba. The input is flattened and input into bidirectional Mamba after fused with Illumination. }
	\label{flowchart} 
\end{figure*}

To this end, this paper, motivated by the efficient Mamba architecture  \cite{zhu2024vision}, designs a 
Mamba-driven deep Retinex unfolding network for low-light enhancement (LLEMamba) that implements Mamba to relight the reflectance and illumination parts for visual guarantee and sticks to the Retinex theory to optimize the closed-form solution of each subproblem for interpretability maintenance. Firstly, the proposed LLEMamba performs deep unfolding on the iteration optimization process of the Retinex model to obtain reflectance and illumination images with mathematical meaning as shown in Fig. \ref{flowchart}.
Secondly, LLEMamba introduces Relighting-Guided Mamba to relight the reflectance, which adopts a novel Illumination-Fused Bidirectional Mamba (IFBM).
Thirdly, the Mamba-driven network utilizes the linear complexity of State Space Model to enhance the Retinex unfolding network efficiently. Additionally, further experiments demonstrate the effectiveness and efficiency of our LLEMamba.

Our main contributions are summarized as follows.

\begin{enumerate}
\item We propose the relighting-guided Mambda with deep unfolding network (LLEMamba), building the first Mamba-based method for low-light enhancement to our knowledge, which could achieve a double win across theoretical interpretability and reconstruction quality.

\item We fuse Mamba into the iterative optimization process of Retinex model with lower computational complexity, which could boost the multiple iterations of deep unfolding networks for stable closed-form solutions.

\item We design a more accurate alternating iterative solution method based on ADMM. Numerous experiments show that our proposed models outperform the state-of-the-art methods on various benchmark datasets. 

\end{enumerate}

\section{Related work}

\textbf{Retinex-based low-light image enhancement: }
Retinex theory, recognized for its role in achieving lightness and color constancy in computer vision \cite{jobson1997properties}, is instrumental in low-light image enhancement. It enables the decomposition of an image into its reflectance and illumination components \cite{li2018structure,lin2022low}. By enhancing the illumination component, Retinex-based models aim to simulate brighter lighting conditions, thereby improving visibility and image quality \cite{guo2016lime,wei2018deep}. To address the challenges (i.e., inevitable noise and poor visibility) in low-light image enhancement, sophisticated neural networks have been developed to tackle the illumination and reflectance components separately. These networks are designed to adjust the light levels and remove degradations caused by poor lighting conditions \cite{zhang2021beyond}.  Furthermore, there has been progress in creating simple CNN-based networks that allow for user-specific illumination adjustment  \cite{zhang2019kindling, wu2022uretinex}. Despite these advancements, current models still face limitations, particularly in accurately recovering the reflectance component. They often struggle with noise and color distortion issues. Consequently, there is a pressing demand for the development of more efficient networks that can effectively enhance illumination and recover reflectance.


\textbf{Vision Mamba:}
The Mamba architecture, initially introduced for its efficiency in training and inference compared to Transformers and Recurrent Neural Networks \cite{gu2023mamba}, has been increasingly adopted in various vision applications, such as image dehazing \cite{zheng2024u}, video understanding \cite{li2024videomamba}, and biomedical image segmentation \cite{ma2024u}. Particularly, a pivotal component of the Mamba framework, the scanning mechanism, is not only instrumental in enhancing efficiency but also in providing essential contextual information for visual tasks \cite{zhang2024survey}. In order to adeptly handle 2D or 3D data, innovative designs of the scanning mechanism and the architecture of Mamba blocks are critical for effectively managing multi-dimensional inputs. For example, bidirectional scanning has been effectively employed in applications such as image classification and semantic segmentation \cite{wang2024large,zhu2024vision}. 
Additionally, a multi-path scanning strategy has been proposed within the frequency domain \cite{zhen2024freqmamba}, with the Frequency Band Scanning Mamba offering a new perspective on 2D modeling. 
Beyond the pure Mamba architectures, hybrid models combining Mamba with other architectural paradigms, such as CNNs \cite{chen2024res} and Transformers \cite{zubic2024state}, are currently being developed. Despite these advances, the potential of Mamba architectures in addressing low-light image enhancement problem, remains largely uncharted and presents an opportunity for further research.


\section{Method}

In this section, we introduce the proposed LLEMamba that consists of three parts, i.e., Retinex-based deep unfolding network, relighting-guided Mamba, and illumination-fused Bidirectional Mamba, as shown in Fig. \ref{flowchart}.

\subsection{Retinex-based deep unfolding network}
Following the Retinex theory, images can be decomposed into reflectance with physical color information and Illumination with the degree of lighting.
Then, given an input image $\mathbf{I} \in \mathbb{R}^{H \times W \times 3}$, it can be formulated as the multiplication of Reflectance $\mathbf{R}\in \mathbb{R}^{H \times W \times 3}$ and Illumination $\mathbf{L}\in \mathbb{R}^{H \times W \times 3}$:
\begin{equation} \label{Retinex}
\begin{array}{lr}  
\mathbf{I} = \mathbf{R} \circ \mathbf{L},
\end{array}
\end{equation}
where $\circ$ represents the pixel-wise multiplication.
$H$ and $W$ denote the spatial dimensions of the original low-light image.

It is known that separating $\mathbf{R}$ and $\mathbf{L}$ from $\mathbf{I}$ is an ill-posed problem.
To address the above ill-posedness issue, several hand-crafted priors have been designed, such as low-rankness for noise suppression \cite{ren2020lr3m}, total variation for noise level map learning \cite{zheng2021adaptive}.
Although these hand-crafted priors have clear mathematical meanings, they have limited reconstruction performance and require manual adjustment for different scenarios. To relight the two decomposed components, we add deep priors for both the reflectance layer and illumination layer as $f(\cdot)$ and $g(\cdot)$. Unlike URetinex-Net \cite{wu2022uretinex}, we design Mamba networks to replace $f(\cdot)$ and $g(\cdot)$. 
The objective function of minimization of the Eq. \eqref{Retinex} can be formulated as: 
\begin{equation} \label{eq2}
\begin{array}{lr}  
    \mathop{\min}\limits_{\mathbf{R},\mathbf{L}} 
    \|\mathbf{R} \circ \mathbf{L}-\mathbf{I}\|_{F}^2 
    + \lambda f(\mathbf{R}) + \gamma g(\mathbf{L}). 
\end{array}
\end{equation}
By introducing two auxiliary variables, i.e., $\mathbf{R}=\mathbf{P}$ and $\mathbf{L}=\mathbf{Q}$, the corresponding Lagrangian function of Eq. \eqref{eq2} can be written as:
\begin{equation} \label{eq3}
\begin{array}{lr}  
    \mathcal{L}=  
    \|\mathbf{R} \circ \mathbf{L}-\mathbf{I}\|_{F}^2 

    + \lambda f(\mathbf{P}) + \gamma g(\mathbf{Q})

    + \frac{\mu}{2}\|\mathbf{R} - \mathbf{P} + \frac{\mathbf{Y_1}}{\mu}\|_F^2 +\frac{\mu}{2}\|\mathbf{L} - \mathbf{Q} +\frac{\mathbf{Y_2}}{\mu}\|_F^2,
\end{array}
\end{equation}
where $\mathbf{Y_2}$ and $\mathbf{Y_1}$ represent the Lagrangian Multipliers.
We employ the Alternative Directional Method of Multipliers (ADMM) to decompose the optimization problem into subproblems with respect to $\mathbf{R}$, $\mathbf{L}$, $\mathbf{P}$, and $\mathbf{Q}$.

\textbf{$\mathbf{R}$-subproblem: }
We update $\mathbf{R}$ by fixing other variables as
\begin{equation} \label{sub-R}
\begin{array}{lr}  
    \mathbf{R}^{(k+1)} =  \mathop{\arg\min}\limits_{\mathbf{R}} 
    \|\mathbf{R} \circ \mathbf{L}^{(k)}-\mathbf{I}\|_{F}^2

    + \frac{\mu^{(k)}}{2}\|\mathbf{R} - \mathbf{P}^{(k)} + \frac{\mathbf{Y_1}^{(k)}}{\mu^{(k)}}\|_F^2.
\end{array}
\end{equation}
We take the derivative with respect to $\mathbf{R}$ and take it to zero.
The closed-form solution of Eq. \eqref{sub-R} is:
\begin{equation} \label{solution-R}
\begin{array}{lr}  
    \mathbf{R}^{(k+1)} = (2 (\mathbf{I} \circ \mathbf{L}^{(k)}) + \mu^{(k)} \mathbf{P}^{(k)}-\mathbf{Y}_1^{(k)})(2 (\mathbf{L}^{(k)} \circ \mathbf{L}^{(k)}) + \mu^{(k)}\mathbf{D})^{-1}.
\end{array}
\end{equation}
where $\mathbf{D} \in \mathbb{R}^{H \times W \times 3}$ is concatenated by three identity matrices along the mode 3.

\textbf{$\mathbf{L}$-subproblem: }
We update $\mathbf{L}$ by fixing other variables as
\begin{equation} \label{sub-L}
\begin{array}{lr}  
    \mathbf{L}^{(k+1)} = \mathop{\arg\min}\limits_{\mathbf{L}} \|\mathbf{R}^{(k+1)} \circ \mathbf{L} - \mathbf{I}\|_F^2 + \frac{\mu^{(k)}}{2} \|\mathbf{L} - \mathbf{Q}^{(k)} + \frac{\mathbf{Y}_2^{(k)}}{\mu^{(k)}}\|_F^2.
\end{array}
\end{equation}
The closed-form solution of Eq. \eqref{sub-L} is given by
\begin{equation} \label{solution-L}
\begin{array}{lr}  
    \mathbf{L}^{(k+1)} = 
(2 (\mathbf{I} \circ \mathbf{R}^{(k+1)}) + \mu^{(k)} \mathbf{Q}^{(k)}-\mathbf{Y}_2^{(k)})(2 (\mathbf{R}^{(k+1)} \circ \mathbf{R}^{(k+1)}) + \mu^{(k)}\mathbf{D})^{-1}.
\end{array}
\end{equation}

\textbf{$\mathbf{P}$-subproblem: }
We update $\mathbf{P}$ by fixing other variables as
\begin{equation} \label{sub-P01}
\begin{array}{lr}  
    \mathbf{P}^{(k+1)}=  
    \mathop{\arg\min}\limits_{\mathbf{P}}
    \lambda f(\mathbf{P}) 
    + \frac{\mu^{(k)}}{2}\|\mathbf{R}^{(k+1)} - \mathbf{P} + \frac{\mathbf{Y_1}^{(k)}}{\mu^{(k)}}\|_F^2.
\end{array}
\end{equation}
By simple algebra, we can obtain
\begin{equation} \label{sub-P02}
\begin{array}{lr}  
    \mathbf{P}^{(k+1)}=  
    \mathop{\arg\min}\limits_{\mathbf{P}}
     f(\mathbf{P}) 
    + \frac{1}{(2\sqrt{\lambda/\mu^{(k)}})^2}\|\mathbf{P} - \mathbf{M}^{(k+1)}\|_F^2,
\end{array}
\end{equation}
where $\mathbf{M}^{(k+1)}=\mathbf{R}^{(k+1)} + \frac{\mathbf{Y_1}^{(k)}}{\mu^{(k)}}$ could be viewed as the noisy image. Thus, the clean image $\mathbf{P}^{(k+1)}$ could be perceived from the noisy image $\mathbf{M}^{(k+1)}$ referring to noise $\alpha$, where $\alpha=\sqrt{\lambda/\mu^{(k)}}$. According to   \cite{zhu2023denoising}, we can also derive Eq.~\eqref{sub-P02} as
\begin{equation} \label{sub-P03}
\begin{array}{lr}  
    \mathbf{P}^{(k+1)}=  
    \mathbf{M}^{(k+1)}+\alpha^{2}f_{\Theta} (\mathbf{M}^{(k+1)}),
\end{array}
\end{equation}
where $f_{\Theta}$ represents the relighting-guided Mamba network and $\Theta$ denotes its network parameter. Compared with URetinex-Net \cite{wu2022uretinex}, we obtain a more accurate $\mathbf{P}$ by embedding the parameters $\lambda$ and $\mu$ into the closed-form solution.

\textbf{$\mathbf{Q}$-subproblem: } We update $\mathbf{Q}$ by fixing other variables as
\begin{equation} \label{sub-Q01}
\begin{array}{lr}  
    \mathbf{Q}^{(k+1)}=  
    \mathop{\arg\min}\limits_{\mathbf{Q}}
    \gamma g(\mathbf{Q}^{(k+1)}) 
    + \frac{\mu^{(k)}}{2}\|\mathbf{L}^{(k+1)} - \mathbf{Q}^{(k+1)} + \frac{\mathbf{Y_2}^{(k)}}{\mu^{(k)}}\|_F^2.
\end{array}
\end{equation}
We can simply improve Eq.~\eqref{sub-Q01} as
\begin{equation} \label{sub-Q02}
\begin{array}{lr}  
    \mathbf{Q}^{(k+1)}=  
    \mathop{\arg\min}\limits_{\mathbf{Q}}
     f(\mathbf{Q}^{(k+1)}) 
    + \frac{1}{(2\sqrt{\gamma/\mu^{(k)}})^2}\|\mathbf{Q}^{(k+1)} - \mathbf{N}^{(k+1)}\|_F^2.
\end{array}
\end{equation}
Similarly, $\mathbf{N}^{(k+1)}=\mathbf{L}^{(k+1)} + \frac{\mathbf{Y_2}^{(k)}}{\mu^{(k)}}$ is treated as the noisy image, while the clean image $\mathbf{Q}^{(k+1)}$ can be recovered from the noisy image $\mathbf{N}^{(k+1)}$ with noise $\delta$. Let $\delta=\sqrt{\lambda/\mu^{(k)}}$   \cite{zhu2023denoising}, Eq.~\eqref{sub-Q02} can be reformulated as
\begin{equation} \label{sub-Q03}
\begin{array}{lr}  
    \mathbf{Q}^{(k+1)}=  
    \mathbf{N}^{(k+1)}+\delta^{2}g_{\Theta} (\mathbf{N}^{(k+1)}),
\end{array}
\end{equation}
where $g_{\Theta}$ represents a Mamba block in Fig. \ref{flowchart} (a).

\textbf{Multipliers: } We update $\mathbf{Y}_1$ and $\mathbf{Y}_2$ as
\begin{equation} \label{sub-Y1}
\begin{array}{lr}  
    \mathbf{Y}_1^{(k+1)}=\mathbf{Y}_1^{(k)} + \mu^{(k)}( \mathbf{R}^{(k+1)}-\mathbf{P}^{(k+1)}).
\end{array}
\end{equation}
\begin{equation} \label{sub-Y2}
\begin{array}{lr}  
    \mathbf{Y}_2^{(k+1)}=\mathbf{Y}_2^{(k)} + \mu^{(k)}( \mathbf{L}^{(k+1)}-\mathbf{Q}^{(k+1)}).
\end{array}
\end{equation}

Finally, the above iteration optimization process can be embedded into 
an unfolding network, as shown in Fig. \ref{flowchart} (a). Based on the Retinex unfolding algorithm, our model carries out decomposition through 4 convolution layers and obtains the initial illumination $\mathbf{L}$ and reflectance $\mathbf{R}$. 
Then, our model conducts deep unfolding optimization on illumination layer $\mathbf{L}$ 
and reflectance layer $\mathbf{R}$ iteratively.

\subsection{Relighting-guided Mamba}
After the decomposition from the last section, the decomposed input image is usually corrupted with color distortion, noise, and over-exposure.
In this paper, we define the process of enhancing and restoring the decomposed layers as Relight, which attempts to restore the physical and illumination details of the overall image after the first decomposition phase.

The existing deep unfolding model on Retinex  \cite{wu2022uretinex} utilized the CNNs as backbone networks, which has limitations for global visual context learning ability.
Although the Transformer-based Retinex network is proposed with promising performance  \cite{cai2023retinexformer}, the GPU memory limit obstructs the iteration of a deep unfolding network.
Therefore, we utilize the pure Mamba as our backbone network to achieve a trade-off memory efficiency and relight effect as shown in Fig. \ref{flowchart} (b).

We adopt a U-shaped network  \cite{cai2023retinexformer} with pure Mamba as our backbone network to restore the reflectance $\mathbf{R}$ and utilize a vanilla mamba to extract illumination feature in $\mathbf{L}$. 
With noise and color distortion, reflectance $\mathbf{R}$ plays a critical role in Relighting.
Given the input decomposed reflectance $\mathbf{R}$, in the encoder of the U-shape network,
$\mathbf{R}$ is passed through a $3\times3$ convolution layer,  an illumination-fused bidirectional Mamba block, a $4\times4$ convolution layer which generates features $\mathbf{F}_i \in \mathbb{R}^{{\frac{H}{2^i}}\times{{\frac{W}{2^i}}\times{2^{i}C}}}$, and then another illumination-fused bidirectional Mamba blocks.
After the downsampling branch, the $2\times2$ deconvolution layer is employed to upscale the features.
Additionally, skip connections are exploited to reduce the loss during downsampling.
In the end, the network outputs a $H\times W \times3$ image with restoration.

\subsection{Illumination-fused Bidirectional Mamba}
\begin{figure*}
	\centering
        \includegraphics[width=1\linewidth]{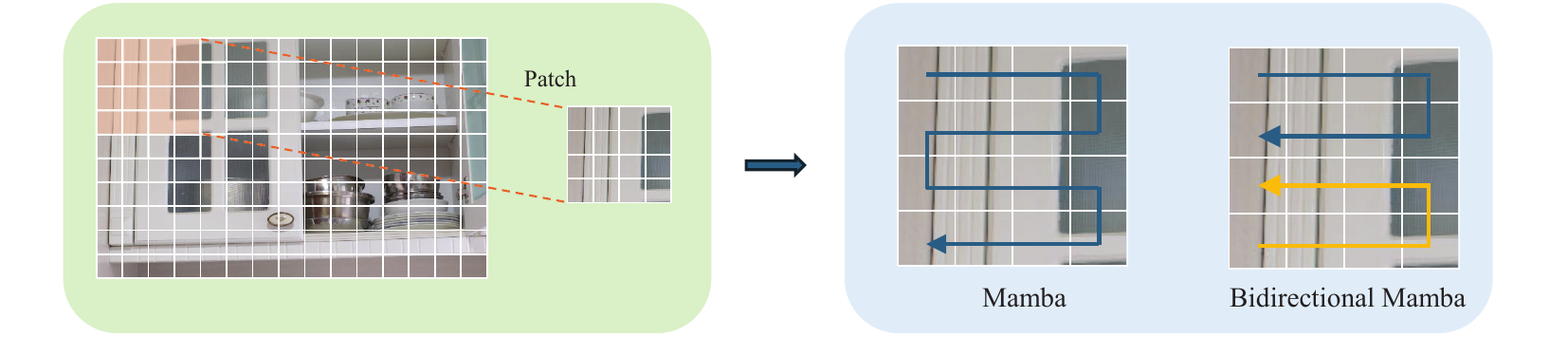}
	\caption{Scanning order of Mamba and Bidirectional Mamba. A bidirectional scan mechanism gathers global color context more accurately.}
	\label{mamba} 
\end{figure*}

According to Eq. \ref{sub-P03}, both illumination $\mathbf{L}$ and reflectance $\mathbf{R}$ are needed for the solution of $\mathbf{R}$.
In addition, the color distortion in a dark area is highly correlated with the illumination condition.
Thus, We take the illumination features into consideration for the Mamba model design as shown in Fig \ref{flowchart} (c).
We fuse the flattened illumination $\mathbf{L}$ and reflectance $\mathbf{R}$ firstly in the illumination-fused bidirectional Mamba (IFBMamba).
The input reflectance $\mathbf{R} \in \mathbb{R}^{H\times W \times C}$
is reshaped into tokens of $\mathbf{R}_p \in \mathbb{R}^{J \times (P^2\dot C)}$,
where $H,W$ is the size of input, $C$ is the number of channels, and $P$ is the size of patches.
$\mathbf{R}_p$ is linearly projected into vectors as follows:
\begin{equation} \label{Mamba}
\begin{array}{lr}  
    \mathbf{T}_0 = [\mathbf{t}_{cls};\mathbf{t}_{p}^{1}\mathbf{W};\mathbf{t}_{p}^{2}\mathbf{W};...;;\mathbf{t}_{p}^{J}\mathbf{W}],
\end{array}
\end{equation}
where $\mathbf{t}_{p}^{J}$ is the $j$-th patch of $\mathbf{t}$, $\mathbf{W}$ is the learnable projection matrix.
To efficiently learn the global color information of these vectors, we adopt a bidirectional Mamba to extract the feature of reflectance instead of a vanilla Mamba.
In the right part of Fig. \ref{mamba}, the sequence of patches goes through a forward-scanning Mamba and a backward-scanning Mamba.
Vanilla Mamba will forget previous information when the scanned length grows longer while bidirectional Mamba can prevent excessive focus on local patches and model the global patches via forward and backwared scanning.

\section{Experiments}

\subsection{Experiment Settings}
\textbf{Datasets.}
LOL-v1  \cite{wei2018deep} is captured in real scenes with different exposure conditions and ISO, which contains 485 low-light and normal image pairs as the training set and 15 low-light and normal image pairs as the test set.
LOL-v2-Real  \cite{yang2021sparse} is the second version of the LOL dataset, with 689:100 training sets and test sets.
LOL-v2-Syn  \cite{yang2021sparse} contains 900 low-light and normal-light image pairs, which are synthesized from captured images by analyzing the distribution of luminance channels of dark and normal images.
SID   \cite{chen2019seeing} is captured by the Sony $\alpha$7S II camera. It contains 2697 short-exposure and long-exposure RAW image pairs, which are taken in sufficient wide dynamic low-light scenes.

\textbf{Evaluation Metrics.}
PSNR \cite{wang2013naturalness} is utilized to measure the quality of images.
SSIM  \cite{wang2013naturalness} is a method used to measure the similarity between two images.
LPIPS  \cite{zhang2018unreasonable} calculates the distance between two images in high-level features.

\textbf{Compared Methods.}
Traditional models: 
LIME  \cite{guo2016lime}, 
MF  \cite{fu2016fusion}, 
NPE  \cite{wang2013naturalness}, 
SRIE  \cite{fu2016weighted} and learning-based models: 
RetinexNet  \cite{wei2018deep}, 
DeepUPE \cite{wang2019underexposed}, 
DRBN \cite{yang2021band}, 
Restormer \cite{zamir2022restormer}, 
MIRNet \cite{zamir2020learning},
Kind  \cite{zhang2019kindling},
Kind++  \cite{zhang2021beyond},
SGM  \cite{yang2021sparse},
RUAS  \cite{liu2021retinex},
SNR-Aware   \cite{xu2022snr},
Bread  \cite{guo2023low},
URetinex-Net \cite{wu2022uretinex},
FourLLIE  \cite{wang2023fourllie}, 
GSAD \cite{hou2024global},
Retinexformer  \cite{cai2023retinexformer}.

\begin{figure*}
	\centering
        \includegraphics[width=1\linewidth]{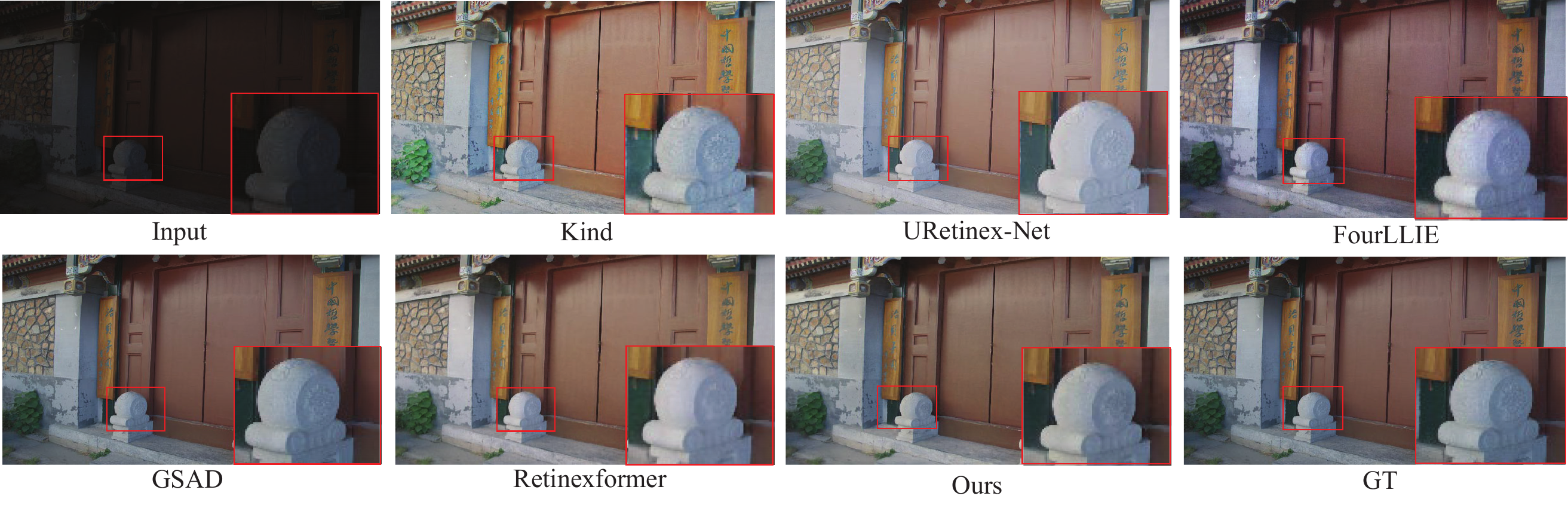}
	\caption{Visual comparison on LOL-v2\cite{wei2018deep} dataset. 
The brightness contrast between objects is well preserved in LLEMamba.}
	\label{visual1} 
\end{figure*}

\textbf{Implementation Details.}
The Initialization and Relight are trained separately, where the batch sizes are set to 4. 
Adam optimizer is utilized   \cite{kingma2014adam} with $\beta_1$ = 0.9 and $\beta_2$=0.999. 
In the relight module, it is trained for $1.5\times10^5$ iterations with initial learning rate = $2\times10^{-4}$ and decreased to $1\times10^{-6}$.  
The initial penalty parameters $\lambda$ and $\gamma$ are set 0.1 and 0.05 respectively Eq. \eqref{eq3}.
All the experiments are conducted on the NVIDIA GeForce RTX 4090
GPU with PyTorch framework.
\begin{table*}[h]\scriptsize

\centering
\caption{Quantitative Comparison on the LOL-v1, LOL-v2-Real, and LOL-v2-Syn Dataset. The highest result is in \textcolor{red}{red} and the second highest result is in \textcolor{blue}{blue}.}
\label{quan1}
\begin{tabular}{lclcccccccc}
    \toprule
    \multirow{2}{*}{Methods} & \multirow{2}{*}{Venue} & \multicolumn{3}{c}{LOL-v1} & \multicolumn{3}{c}{LOL-v2-Real} & \multicolumn{3}{c}{LOL-v2-Syn} \\
    \cmidrule(r){3-5} \cmidrule(r){6-8} \cmidrule(r){9-11}
    & & PSNR & SSIM & LPIPS & PSNR & SSIM & LPIPS & PSNR & SSIM & LPIPS \\
    \midrule
    NPE \cite{wang2013naturalness} & TIP2013 & 16.97 & 0.589 & 0.416 & 17.33 & 0.464 & 0.236 & 16.60 & 0.778 & 0.187 \\
    LIME \cite{guo2016lime}& TIP2016 & 16.76 & 0.578 & 0.418 & 15.24 & 0.419 & 0.220 & 16.88 & 0.758 & 0.204 \\
    MF \cite{fu2016fusion} & Signal Processing & 18.79 & 0.642 & 0.407 & 18.72 & 0.509 & 0.240 & 17.50 & 0.774 & 0.208 \\
    SRIE \cite{fu2016weighted} & CVPR2016 & 11.86 & 0.498 & 0.423 & 14.45 & 0.524 & 0.216 & 14.50 & 0.664 & 0.248 \\
    RetinexNet \cite{wei2018deep} & BMVC2018 & 13.10 & 0.429 & 0.325 & 16.08 & 0.656 & 0.236 & 18.28 & 0.774 & 0.234 \\
    Kind \cite{zhang2019kindling} & MM2019 & 20.87 & 0.802 & 0.240 & 20.01 & 0.641 & 0.081 & 22.01 & 0.904 & 0.273 \\
    Kind++ \cite{zhang2021beyond} & IJCV2021 & 21.30 & 0.823 & 0.262 & 20.59 & 0.829 & 0.088 & 21.07 & 0.881 & 0.267 \\
    SGM \cite{yang2021sparse} & TIP2021 & 20.06 & 0.816 & 0.215 & 20.06 & 0.816 & 0.073 & 22.05 & 0.909 & 0.484 \\
    RUAS \cite{liu2021retinex} & CVPR2021 & 18.23 & 0.717 & 0.270 & 18.37 & 0.723 & 0.181 & 16.55 & 0.652 & 0.579 \\
    SNR-Aware  \cite{xu2022snr} & CVPR2022 & \textcolor{blue}{23.61} & \textcolor{blue}{0.842} & 0.262 & {21.48} & \textcolor{blue}{0.848} & 0.074 & 24.13 & \textcolor{red}{0.927} & 0.342 \\
    URetinex-Net \cite{wu2022uretinex} & CVPR2022 & 21.33 & 0.835 & \textcolor{blue}{0.050} & 21.54 & 0.801 & \textcolor{red}{0.044} & \textcolor{blue}{24.23} & {0.897} & \textcolor{red}{0.128} \\
    Bread \cite{guo2023low} & IJCV2023 & 22.92 & 0.836 & 0.160 & 20.83 & {0.822} & 0.095 & 17.63 & 0.838 & 0.168 \\
    FourLLIE \cite{wang2023fourllie} & MM2023 & 19.34 & 0.757 & 0.338 & 19.09 & 0.770 & 0.380 & 16.85 & 0.852 & 0.423 \\
        NerRCo \cite{yang2023implicit} & ICCV2023 & 22.95& 0.815 & 0.338 & \textcolor{red}{24.17} & 0.813 & 0.311 & 19.07 & 0.714 & 0.463 \\
    \midrule
    LLEMamba & Ours & \textcolor{red}{23.71 }& \textcolor{red}{0.901} & \textcolor{red}{0.046} & \textcolor{blue}{22.20} & \textcolor{red}{0.901} & \textcolor{blue}{0.054} & \textcolor{red}{24.27} & \textcolor{blue}{0.914} & \textcolor{blue}{0.142} \\
    \bottomrule
  \end{tabular}
\end{table*}
\subsection{Experimental Results}
\textbf{Quantitative Analysis.}
In the quantitative comparison shown in Table \ref{quan1}, our proposed LLEMamba achieves the highest results on average with evaluation metrics of PSNR, SSIM, and LPIPS.
In the LOL-v1 dataset, we first make comparison with SOTA Retinex-based methods, i.e. RetinexNet \cite{wei2018deep}, Kind \cite{zhang2019kindling}, Kind++ \cite{zhang2021beyond}, RUAS \cite{liu2021retinex}, URetinex-Net \cite{wu2022uretinex}, with  10.61, 2.84, 2.41, 5.48, and 2.38dB improvements in PSNR respectively.
Especially for the deep unfolding based network URetinex-Net, our proposed LLEMamba has shown superior image quality with higher PSNR, which shows the relight effect of the proposed LLEMamba.
In the LOL-v2-Real dataset, LLEMamba achieves the highest in PSNR, SSIM, and LPIPS.

LLEMamba also outperforms other methods on average metrics in the LOL-v2-Syn dataset. 
Meanwhile, URetinex-Net\cite{wu2022uretinex} has the highest result in a few metrics, which comes from its overfitting on the LOL-v2 dataset.
Specifically, the deep unfolding network preserves the physical information in both the illumination and reflectance layers, and the proposed IFBMamba boosts the long-range sequence learning with superb denoising and restoration effects.
Generally, the superb performance on PSNR has shown the denoising effect of LLEMamba, and the excellent scores on SSIM and LPIPS prove its excellent global visual context modeling.

In Table \ref{SID}, our proposed LLEMamba has surpassed the existing SOTA methods in all metrics. 
Compared with advanced Retinexformer\cite{cai2023retinexformer}, LLEMamba achieves higher PSNR and SSIM in the SID dataset which involves both indoor and outdoor real-scene low-light images with stable long-range dependency learning.
As for the model parameters, our LLEMamba has decreased 6.25\% parameters compared with transformer-based Retinexformer \cite{cai2023retinexformer}, which could support deep unfolding networks in low-level vision tasks. 
The SID \cite{chen2019seeing} dataset involves dark images indoors and outdoors with different noise levels, thus LLEMamba proves robust learning ability in different low-light scenes.

\begin{figure*}
	\centering
        \includegraphics[width=1\linewidth]{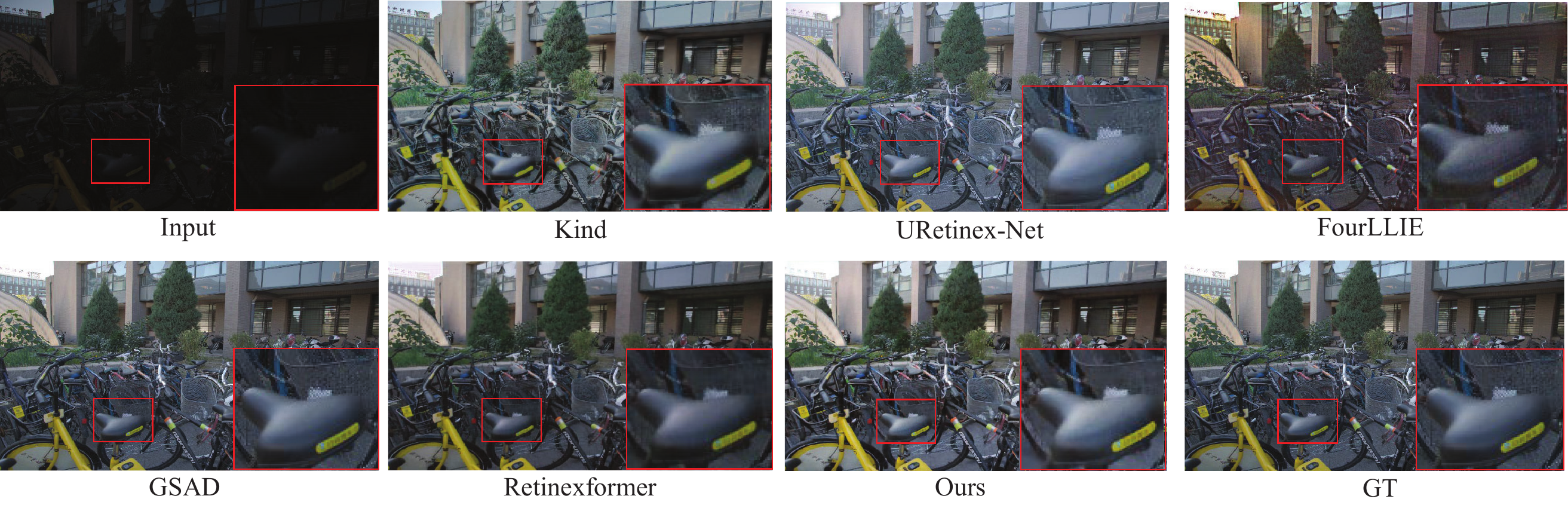}
	\caption{Visual comparison on LOL-v2\cite{wei2018deep} dataset. LLEMamba accurately captures the texture details of objects.}
	\label{visual2} 
\end{figure*}

\textbf{Qualitative Analysis.}
To facilitate an intuitive comparison, we make visual results comparison with classic Retinex-based methods: Kind \cite{zhang2019kindling}, URetinex-Net \cite{wu2022uretinex}, FourLLIE \cite{wang2023fourllie}, and Retinexformer \cite{cai2023retinexformer}.
Compared with Kind \cite{zhang2019kindling} and FourLLIE \cite{wang2023fourllie}, the proposed LLEMamba can capture color information more precisely.
Additionally, LLEMamba prevents over-exposure in the light area compared with URetinex \cite{wu2022uretinex} in the background of Fig. \ref{visual2}.

\begin{figure*}
	\centering
        \includegraphics[width=1\linewidth]{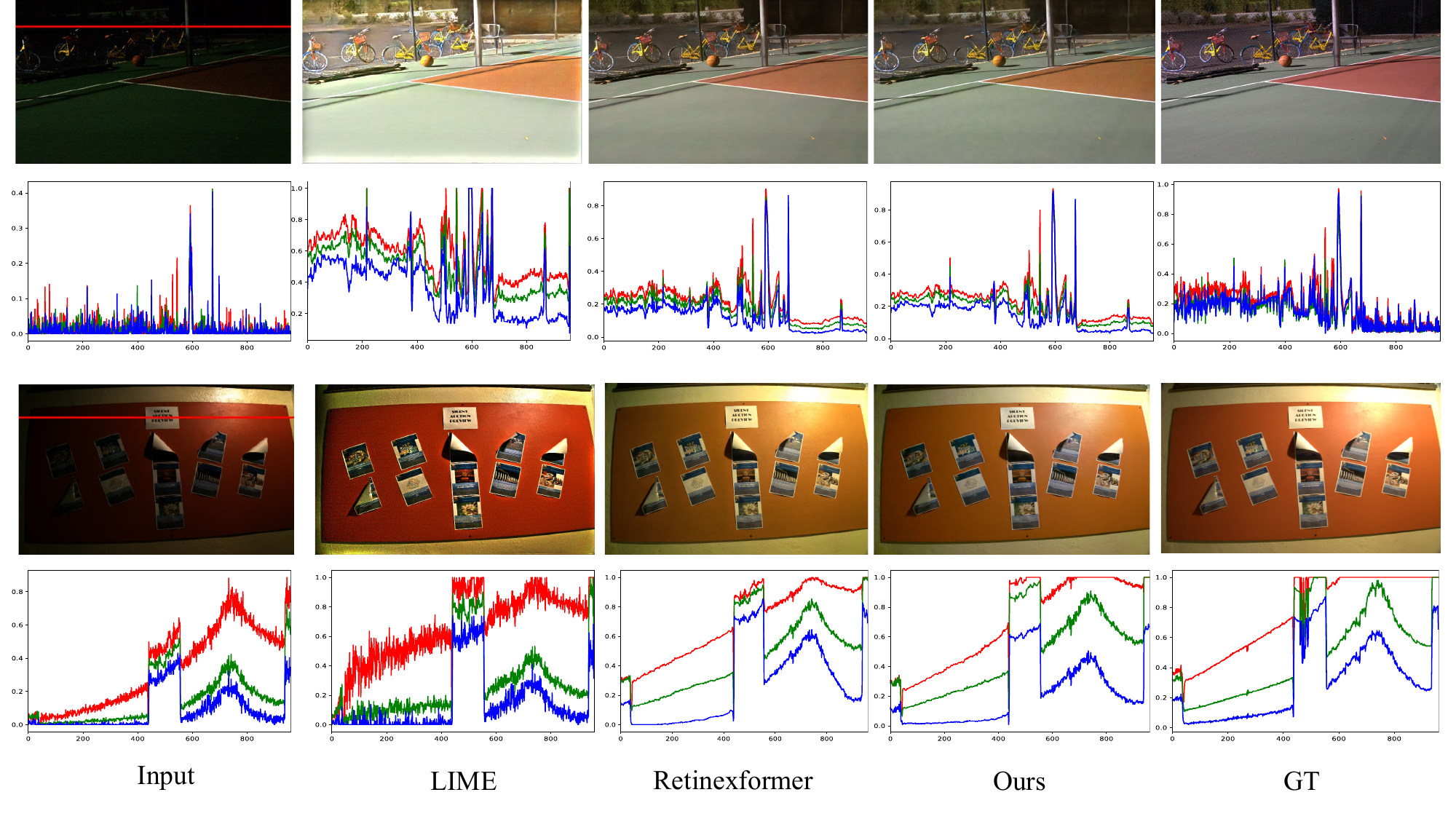}
	\caption{Visual comparison on SID dataset. The 1-D RGB signals in each image at the red line markers in (a) are represented separately in red, green, and blue. The horizontal axis represents the pixel index, while the vertical axis represents the corresponding pixel value.}
	\label{rgb} 
\end{figure*}

In the SID dataset which contains more noise in the input image, traditional LIME has a severe over-exposure problem and Retinexformer may fail to learn the precise color in reflectance of the bicycles in Fig. \ref{rgb}.
The 1-D RGB plots depict the pixel values in the images, from which we can obtain the following information: the max value in the curve and the sharp changes in the curve.
Unlike the curves of other methods with mismatched max values, our proposed LLEMamba
prevent over-exposure at the peak of the curves.
Moreover, the sharp change tendency of LLEMamba is fitted with the curves of GT precisely.
Thus, LLEMamba outperforms other SOTA algorithms in multiple dark scenes.

\begin{table*}[h]\tiny
\centering
\caption{Quantitative Comparison with Different Methods on SID dataset. The highest result is in \textcolor{red}{red} and the second highest result is in \textcolor{blue}{blue}.}
\label{SID}
\begin{tabularx}{\textwidth}{l|l|*{8}{c|}c}
\toprule
\multicolumn{2}{l|}{} & DeepUPE & {DRBN} & {KinD} & {SGM}  & {Restormer} & {MIRNet} & {SNR-Aware} & {Retinexformer} & {Ours} \\
\midrule
\multirow{2}{*}{SID} & {PSNR} & 17.01 & 19.02 & 18.02 & 18.68 & 22.27 & 20.84 & 22.87 & \textcolor{blue}{24.44} & \textcolor{red}{24.55} \\
& {SSIM} & 0.604 & 0.577 & 0.583 & 0.606 &  0.649 & 0.605 & 0.625 & \textcolor{blue}{0.680} & \textcolor{red}{0.682} \\
\midrule
\multirow{2}{*}{Complexity} & {FLOPS(G)} & 5.86 & 48.61 & 34.99 & 53.26  & 144.25 & 785 & 26.35 & {15.57} & {13.96} \\
& {Parameters(M)} & 1.77& 5.27 & 0.583 & 0.606 & 26.13 &  31.76 & 4.01 & {1.6} & {1.5} \\
\bottomrule
\end{tabularx}
\end{table*}

\subsection{Ablation Study}

\begin{wrapfigure}{r}{0.5\textwidth}
    \centering
    \includegraphics[width=0.48\textwidth]{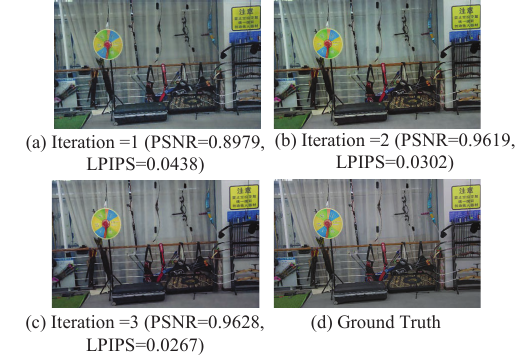}
    \caption{Ablation study of the deep unfolding iterations.}
    \label{iteration} 
\end{wrapfigure}

\textbf{Effect of Deep Unfolding Network.} To explore the relationship between visual performance and deep unfolding iterations, we conduct experiments on the LOL-v2 dataset with iterations = 1, 2, 3. The enhanced images are taken at their corresponding highest PSNR model weight. As shown in Fig. \ref{iteration}, for most images, deeper iteration models better illumination and color information. To achieve a trade-off between training cost and enhancement effect, we suggest selecting iteration = 3 on the LOL-v2 dataset. In large-scale datasets such as SID with limited GPU memory, deeper unfolding results in a decrease in batch size, which deteriorates model convergence.

\begin{minipage}[t]{0.45\textwidth}
\textbf{Effect of Low-Light Enhancement Mamba.}
We analyze the quantitative results of our proposed illumination-fused bidirectional mamba by comparing them with convolutional networks and 
transformer-based model  \cite{cai2023retinexformer}.
The proposed illumination-fused bidirectional mamba in the reflectance $\mathbf{P}$ optimization module is replaced with other networks. 
The implementation is as follows: DnCNN along with SENet in the reflectance enhancement in URetinex-Net \cite{wu2022uretinex},

\end{minipage}%
\hfill
\begin{minipage}[t]{0.5\textwidth}
    \centering
    \scriptsize
    \captionof{table}{Comparison with different backbones in deep unfolding network on the LOL-v1 and LOL-v2-Real Dataset. The best results are boldfaced.}
\begin{tabular}{lcccccc}
    \toprule
    \multirow{2}{*}{Backbone} & \multicolumn{2}{c}{LOL-v1} & \multicolumn{2}{c}{LOL-v2} \\
    \cmidrule(r){2-3} \cmidrule(r){4-5}
     & PSNR & SSIM & PSNR & SSIM \\
    \midrule
    DnCNN + SENet & 21.20 & 0.810 & 20.52 & 0.804 \\
    IGMA & 21.70 & 0.822 & 21.29 & 0.829 \\
    Vanilla Mamba & 20.14 & 0.704 & 19.97 & 0.789 \\
    IFBMamba & \textbf{23.71} & \textbf{0.901} & \textbf{21.80} & \textbf{0.901} \\
    \bottomrule
\end{tabular}
\end{minipage}

 Illumination-guided Multi-attention (IGMA) in Retinexformer \cite{cai2023retinexformer}, and vanilla Mamba block.
The quantitative results on the LOL-v2 dataset suggest the design of our IFBMamba fused with iterative illumination $\mathbf{L}$ is effective with its long-range dependency capture ability.

\section{Conclusion}

Based on Retinex theory, this paper introduces the concept of Relight, which aims to denoise and restore degraded reflectance. To address the relight problem, we present a relighting-guided Mamba embedding within a Retinex unfolding network with mathematical meaning. Through experiments conducted on four widely used low-light benchmarks, we demonstrate the effectiveness of the proposed model from both quantitative and qualitative perspectives.

There are several promising avenues for future research. One option is to develop a more lightweight Mamba block in the backbone network, enabling the use of large-scale datasets for deep unfolding architectures. 
Additionally, exploring other computer vision tasks with the Mamba-driven deep unfolding framework holds great potential for advancing these fields. 
By adapting and extending our framework to these problems, we can leverage the inherent strengths of deep unfolding Mamba in tasks such as semantic segmentation, object detection, and hyperspectral imaging.
\bibliographystyle{unsrt}
\bibliography{main}

\end{document}